\documentclass[letterpaper, 10 pt, conference]{ieeeconf}
\IEEEoverridecommandlockouts
\usepackage{graphicx}
\usepackage{subcaption}
\usepackage[dvipsnames*,svgnames*,transparent]{xcolor}
\usepackage{amsmath}

\usepackage{colortbl}
\usepackage{booktabs}
\usepackage{tabularx}
\usepackage{tabularray}
\usepackage{adjustbox}
\usepackage{multirow}
\usepackage{siunitx}
\usepackage{amsfonts}

\makeatletter
\let\NAT@parse\undefined
\makeatother
\usepackage[hidelinks]{hyperref}
\usepackage[nameinlink,capitalise]{cleveref}
\usepackage{url}

\usepackage{tikz}
\usetikzlibrary{shadows}
\usetikzlibrary{shadows.blur}
\usetikzlibrary{arrows.meta}
\usepackage[skins]{tcolorbox}
\usepackage{float}
\definecolor{Cerulean}{rgb}{0,0,0.95}
\definecolor{LimeGreen}{rgb}{0.15,0.65,0.15}
\definecolor{RoyalBlue}{rgb}{0.17,0.49,0.75}
\definecolor{TableBlue}{rgb}{0.17,0.49,0.75}
\definecolor{Rose}{rgb}{1.0, 0.15, 0.21}
\definecolor{Orange}{rgb}{1.0, 0.5, 0.0}
\definecolor{Gray}{gray}{0.3}
\definecolor{Black}{gray}{0.0}
\definecolor{Purple}{rgb}{0.77,0.12,0.64}

\begin{document}
\title{\LARGE \bf Hard Cases Detection in Motion Prediction by Vision-Language Foundation Models}\author{Yi~Yang$^{1,2}$, Qingwen~Zhang$^{1}$, Kei~Ikemura$^{1}$, Nazre~Batool$^{2}$, John~Folkesson$^{1}$
\thanks{$^{1}$Authors are with the Division of Robotics, Perception, and Learning (RPL), KTH Royal Institute of Technology, Stockholm 114 28, Sweden. (email: yiya@kth.se). 
}
\thanks{$^{2}$Authors are with Research and Development, Scania CV AB, Södertälje 151 87, Sweden.}
}

\maketitle
\begin{abstract}
Addressing hard cases in autonomous driving, such as anomalous road users, extreme weather conditions, and complex traffic interactions, presents significant challenges. To ensure safety, it is crucial to detect and manage these scenarios effectively for autonomous driving systems. 
However, the rarity and high-risk nature of these cases demand extensive, diverse datasets for training robust models. 
Vision-Language Foundation Models (VLMs) have shown remarkable zero-shot capabilities as being trained on extensive datasets.
This work explores the potential of VLMs in detecting hard cases in autonomous driving.
We demonstrate the capability of VLMs such as GPT-4v in detecting hard cases in traffic participant motion prediction on both agent and scenario levels.
We introduce a feasible pipeline where VLMs, fed with sequential image frames with designed prompts, effectively identify challenging agents or scenarios, which are verified by existing prediction models.
Moreover, by taking advantage of this detection of hard cases by VLMs, we further improve the training efficiency of the existing motion prediction pipeline by performing data selection for the training samples suggested by GPT.
We show the effectiveness and feasibility of our pipeline incorporating VLMs with state-of-the-art methods on NuScenes datasets. 
The code is accessible at \textcolor{magenta}{\url{https://github.com/KTH-RPL/Detect_VLM}}.

\end{abstract}

\section{Introduction}
\thispagestyle{plain}
Autonomous driving has witnessed rapid growth due to advancements in deep learning in both academic research and the industry. One of the remaining challenges is how can self-driving cars address various complex and unpredictable scenarios. These include dealing with unusual behaviors from other road users, navigating in extreme weather conditions, responding to emergency traffic situations, managing intricate interactions, and so on. Such scenarios are not only difficult to understand but also pose substantial safety concerns.  They require extensive training data due to their sparsity in the whole dataset and high variability \cite{yang2023prediction}. This data should cover a wide range of these challenging situations to enhance the ability of learning-based autonomous driving systems to respond effectively.

A straightforward but expensive way is to collect more real-world data. Modern datasets are getting bigger and bigger \cite{Argoverse2,waymo,nuplan,alibeigi2023zenseact}, offering a wealth of information.
Additionally, researchers are exploring different approaches such as generating synthetic datasets conditioned on specific needs using generative models \cite{song2023synthetic, yang2020surfelgan, marathe2023wedge}, or reconstructing 3D environments to create new data by manipulating elements like moving or adding road users \cite{tonderski2023neurad,yang2023unisim,kerbl3Dgaussians}.
These techniques expand the size of datasets and can create customized scenarios according to training needs.
However, they tend to be expensive and often require much human intervention.
Incremental learning presents another strategy, where the system continuously adapts by focusing on hard samples where the prediction model has unsatisfied results \cite{mirza2022efficient, yang2022continual}. 
This approach, while more automated, has the weaknesses of lacking interpretability and dependence of the identification of hard samples on the network under training.
This raises the question: is there a more {\bf explainable and independent} method available?

With Large Language Model (LLM) and Vision-Language Foundation Models (VLMs) showing remarkable performance, armed with human-like common sense derived from extensive datasets and large network architectures, the surprising zero-shot capability across varied tasks has quickly gained huge attention \cite{clip, gpt3, touvron2023llama, touvron2023llama2, flamingo, openai2023gpt4}.
As a result, there is a growing interest among researchers to rethink the current autonomous driving pipeline and explore the potential of integrating VLM into it.

There are a number of works across a  range of tasks in the autonomous driving pipeline that use LLM/VLM to enhance various components including perception, prediction, planning, end-to-end learning, etc \cite{cui2024survey, zhou2023vision, ding2023hilm, keysan2023can, yang2024human}. Rather than replacing the current existing pipeline with LLM/VLM, a more realistic and intriguing question emerges: how can current state-of-the-art methods benefit from the integration of these large models? This leads us to revisit the challenge of processing hard cases in driving scenarios and ask the following question: Can VLMs effectively assist in hard case detection?
One of the key strengths of VLMs is their extensive knowledge base, enabling them to comprehend variable images. Moreover, their ability to provide textual explanations can build a more transparent process of detecting such cases. 

Therefore, in this paper, {\bf we explore leveraging VLMs for detecting hard cases at both the agent-level and scene-level, focusing on motion prediction.}
At the agent-level, the goal is to identify road users with unexpected behaviors, which often cause current algorithms to fail resulting in large prediction displacement errors. 
At the scene-level, it is useful to pinpoint challenging scenarios, such as unusual traffic patterns, emergencies, extreme weather conditions, etc.
These scenarios often pose difficulties for existing motion prediction networks.
Therefore, having a pipeline that can recognize and be aware of potential failure is crucial.

\begin{figure*}[htbp]
\centering
\includegraphics[width=0.93\linewidth]{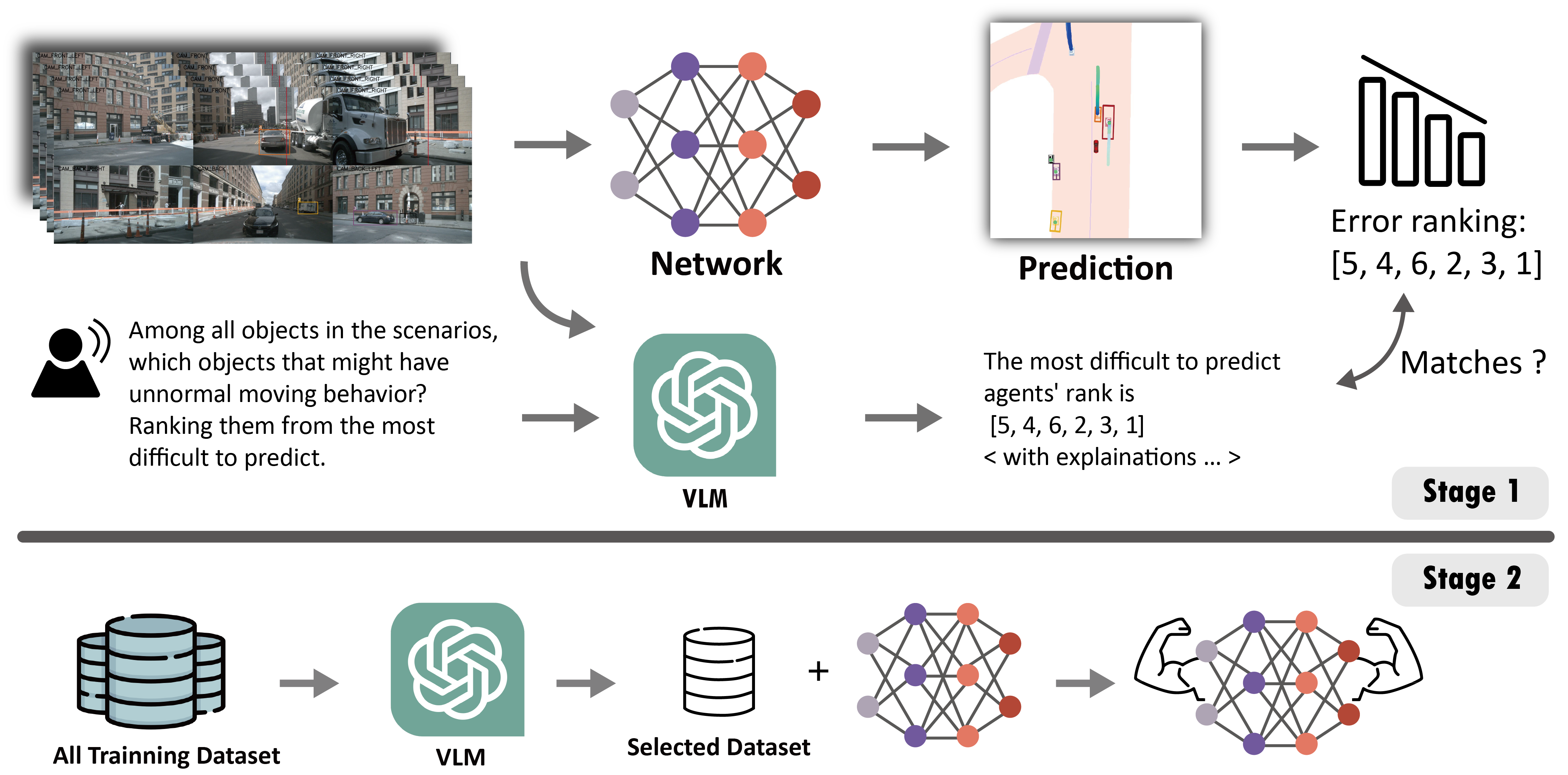}
\caption{\textbf{Two-stages of evaluations.} \textit{Stage 1:} Verify the ability of VLM to detect hard cases, using existing motion prediction results as ground truth. We examine if the VLM's prediction of the most difficult-to-predict agents matches the order based on the highest displacement error in existing motion prediction networks. \textit{Stage 2:} Improve training efficiency by training the network with a smaller subset of data selected by VLM.}
\label{fig:arc}
\vspace{-1.0em}
\end{figure*}

We first demonstrate through our pipeline that VLMs can help effectively detect hard cases in alignment with existing prediction methods. 
Following this, we further explore the rich potential usage of VLMs in detecting hard cases, for instance, data augmentation for difficult scenarios, dataset rebalancing, loss function modulation to give higher weight to hard cases, error analysis, etc.
Here we use data selection as an example to show that the detection of hard cases by VLMs can make the training process more efficient. By selecting proper training samples, thanks to VLMs, training costs are reduced while maintaining comparable performance levels. We present results on the NuScenes dataset \cite{caesar2020nuscenes} to demonstrate the feasibility of integrating VLMs with existing methods. 

We summarize our contributions below:
\begin{itemize}
\item We introduce a feasible pipeline to leverage VLM to detect hard cases in autonomous driving contexts.
\item We verify the detection capability of VLM using the existing prediction networks.
\item We demonstrate through our pipeline that by detecting hard cases, VLM can further facilitate more efficient network training via proper data selection.
\end{itemize}

\section{Related Work}
\subsection{Motion Prediction Using Camera Images}
\label{related_work:motion_pred}
Recently, multi-task and end-to-end approaches for autonomous driving have been brought to great attention \cite{hu2021fiery,liang2020pnpnet,vip3d,uniad,ye2023fusionad}. These methods differ from traditional pipelines as they can directly perform motion prediction, often being jointly optimized with detection and/or tracking, using a single network based on sensor inputs \cite{vip3d, uniad}. 
Hu \textsl{et al.} \cite{hu2021fiery} perform future occupancy prediction in the Bird-Eye View (BEV) space directly from image data. Taking a step forward, Liang \textsl{et al.} \cite{liang2020pnpnet} and Gu \textsl{et al.} \cite{vip3d} incorporate agent-wise information, by jointly optimizing perception, tracking, and motion prediction in an end-to-end manner from sensor data, such as LiDAR readings, and camera images together with HD maps. UniAD \cite{uniad} further includes motion planning into the pipeline and optimized jointly using a query-based approach, by largely promoting the Vision Transformer (ViT) \cite{vit} architecture. Most recently, Ye \textsl{et al.} \cite{ye2023fusionad} propose data fusion in the BEV space prior to the subsequent differentiable modules that perform perception, prediction and planning. End-to-end approaches address the error accumulation and information loss problems in a traditional modular and non-differential pipeline \cite{vip3d, uniad}, thus gaining much interest. 

In this work, we focus on the prediction algorithm using raw images as part of the input. 
Images provide richer information for detecting abnormal or uncommon behaviors and situations compared to bounding boxes alone as typically used in post-perception prediction approaches. 
For instance, images can reveal whether a driver is present in a stopped vehicle. Particularly for pedestrians, useful information such as poses, facial expressions, and gestures can be captured.

\begin{table*}

\centering

\begin{minipage}{\linewidth}
\centering  
\scalebox{0.9}{
\begin{tabular}{l p{15.5cm} }
\toprule
 \multicolumn{2}{l}{\bf VLM prompts design with test images.}  \\
\midrule
a) System & \scriptsize{\texttt{{You are a mature driver behind the wheel. You will see consecutive frames. Each frame has six images captured and put together by the surround view camera in your vehicle. 
The field of view (FOV) of each surround-view camera is approximately 120°. The view angle is written in the images. 
On the right, it is the bird-eye-view image with labeled agents, centered on your vehicle. 
The HD map is also plotted in the bird-eye-view image. You need to answer the following questions:}}}\\
 & \scriptsize{\texttt{\color{Orange}Question 1: \color{Gray}Among all objects, which objects might have abnormal moving behavior that you might need to pay more attention to?
Rank it from the most difficult to predict to the least difficult.}} \\ 
&  \scriptsize{\texttt{\color{Orange}Question 2: \color{Gray} Score the difficulty of prediction from 1 to 10, where 1 is the easiest and 10 is the most difficult. 
If the traffic is very dense and at the crossroads or it is bad weather, you might give it a high score. 
If it is a straight road with few vehicles, it is easy to predict, you might give it a low score. 
If there is an anomalous behavior or situation, you might give it a high score.
}} \\
& \scriptsize{\texttt{Present your answers in this format:}} \\
& \scriptsize{\texttt{\color{LimeGreen}1. The most difficult to predict agents' rank is [A1, A2, A3, A4, A5, A6, A7, A8]. <explanation> Replace A1-A8 with the agent ID number, and make sure the output includes all agents.}} \\
& \scriptsize{\texttt{\color{LimeGreen}2. Overall, the prediction difficulty is <score>. From 1 to 10. <explanation>}} \\
\midrule
b) Assistant & [{\it A few shots are provided to instruct GPT-4V.}] \\
\midrule
c) User  & [{\it Evaluate on test images.}] \\
& \scriptsize{\texttt{\color{Purple}There are 5 different IDs of agents in these images. From ID 1 to ID 5, they are pedestrian, pedestrian, car, motorcycle, and pedestrian.}}  \vspace{-50mm}\\ 
& {
\begin{minipage}{\linewidth}
\begin{figure}[H]
\centering
  \includegraphics[width=\linewidth]{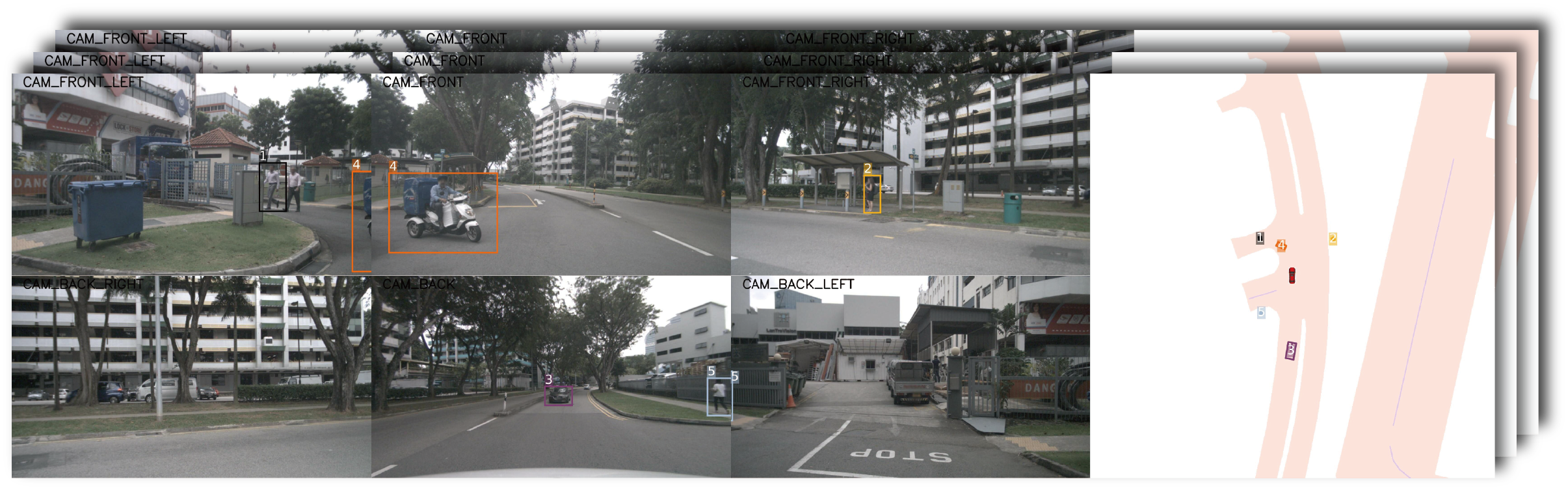}
  \label{fig:enter-label}
\end{figure}
\end{minipage}} \\
GPT-4v~\cite{openai2023gpt4} 
&  \scriptsize{\texttt{\color{RoyalBlue}Output:~\color{Gray}}} \\
& \scriptsize{\texttt{\color{RoyalBlue}1: The most difficult to predict agents' rank is [4, 1, 2, 5, 3]. \color{Black}Explanation: Motorcycle 4 is in motion and could change speed or direction suddenly, making it unpredictable. Pedestrian 1 is close to the crosswalk and might decide to cross the street, which requires attention. Pedestrian 2 is also near the crosswalk and could potentially enter the street. Pedestrian 5 is walking along the sidewalk and seems less likely to enter the street suddenly. Car 3 is parked and not showing any signs of movement, making it the least difficult to predict.}} \\
& \scriptsize{\texttt{\color{RoyalBlue}2: Overall, the prediction difficulty is 5. \color{Black}Explanation: The road is not very busy, and most agents are behaving predictably. However, the presence of pedestrians near the crosswalk and a moving motorcycle increases the level of caution required.}}
\\
\bottomrule
\end{tabular}
}
\vspace{2mm}
\caption{Designed prompts for our approach. a) GPT-4v \cite{openai2023gpt4} is instructed to answer \textcolor{Orange}{two questions} with \textcolor{LimeGreen}{certain format}. b) Few-shot learning: two examples are provided to VLM to learn. c) Given \textcolor{Purple}{inputs} that include 3 frames of consecutive camera images from six different views and their description, GPT-4v \textcolor{RoyalBlue}{outputs the answer} in a required format along with explanations. Here we show a real output from GPT-4v, where it correctly associates most agents with corresponding marks and types, and reasons about their respective states. In addition, descriptive texts are given to reason the generated ranking of agents and difficulty score of the scene, showing superior explainability.}
\label{tab:user_example}  
\end{minipage}
\vspace{-1.0em}
\end{table*}

\subsection{Challenging Cases Estimation in Motion Prediction}
Estimating challenging cases for autonomous driving systems has been the subject of extensive research \cite{shafaei2018uncertainty, mohseni2019predicting, delecki2022we, hecker2018failure, farid2023task}. The difficulty of a driving scenario can be defined with respect to different objectives, including perception \cite{antonante2023task, antonante2023monitoring}, prediction \cite{farid2023task}, planning \cite{hecker2018failure, mohseni2019predicting}, etc. 
In this work, we focus on hard cases in motion prediction. In this regard, Farid \textsl{et al.} \cite{farid2023task} propose a $p$-Quantile Anomaly Detection (QAD) algorithm to perform online failure prediction of non-ego agent trajectory estimation results. The method is task-relevance: it prioritizes and assigns greater weight to a subset of false predictions that are most likely to impact the planning maneuvers of the ego vehicle. Kuhn \textsl{et al.} \cite{kuhn2020introspective} introduce an introspective approach to predict future failure by learning from previous disengagement sequences. Stocco \textsl{et al.} propose ThirdEye \cite{stocco2022thirdeye} that detects unsafe conditions such as adverse weather, lightning, etc., using the attention maps produced by the prediction network during driving. The aforementioned methods employ heuristic rules or network training to estimate uncertainty or identify challenging cases.

\begin{figure*}[htbp]
\centerline{\includegraphics[width=\linewidth]{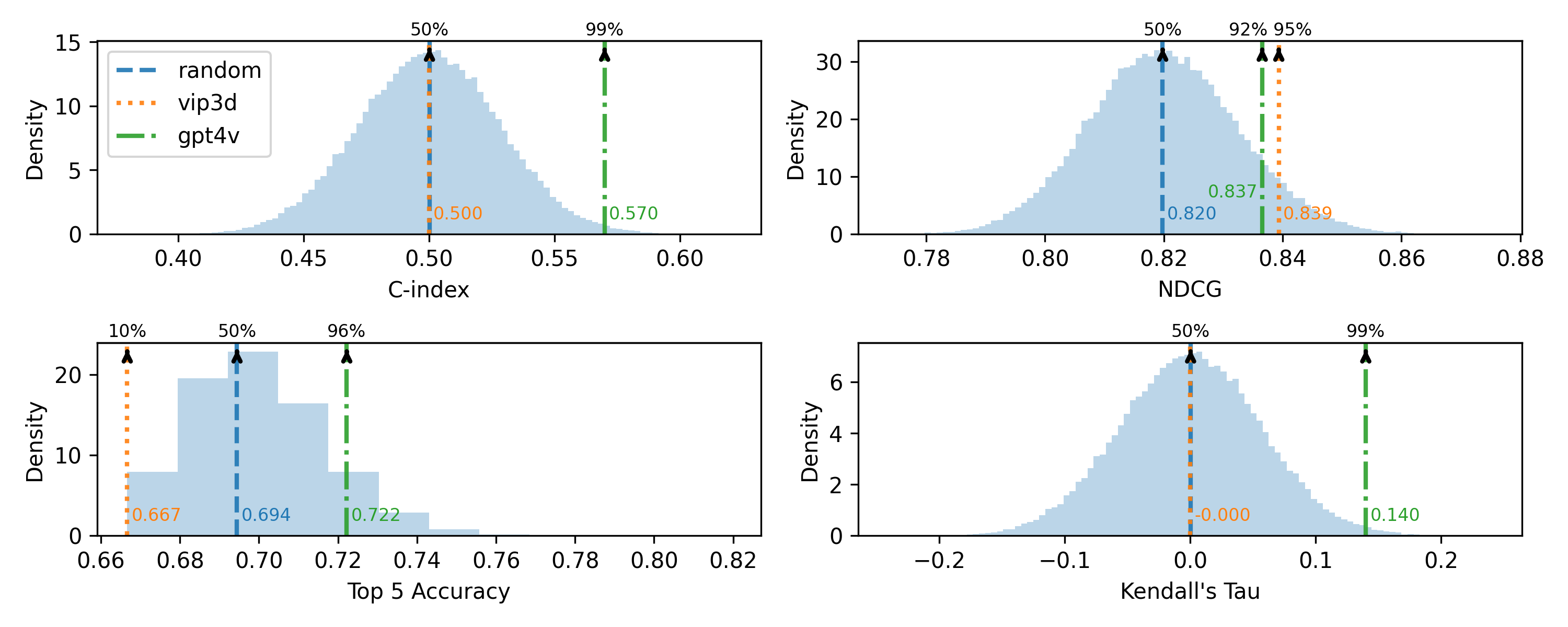}}
\caption{Result of agents ranking according to higher prediction error / difficulty. Using the UniAD \cite{uniad} ranking as ground truth, we compare it with random order, order from ViP3D \cite{vip3d}, and GPT-4v. The evaluation is conducted using four metrics: C-index, NDCG, top-5 accuracy, and Kendall's Tau, where larger values indicate a higher correlation with UniAD order. The x-axis is the metric value. Note that for the random ordering, we conducted 10,000 trials, and the distribution of the results is shown in the blue histogram, with the y-axis representing the probability density / frequency; note that the metric value of random is the mean from all trials. The percentage values above the graph indicate the percentage of random trials that are surpassed by this value (cumulative probability).}
\label{fig:res_}
\vspace{-1.0em} 
\end{figure*}

\subsection{LLMs and VLMs for Autonomous Driving Systems}

Large Language Model (LLM) refers to a specific type of Transformer-based language model featuring extensive numbers, often hundreds of billions, of parameters \cite{cui2024survey}. 
These models exhibit remarkable generalizability after being trained with text prediction objectives using vast amounts of internet data.
Some notable examples of LLMs include GPT-3 \cite{gpt3}, PaLM \cite{chowdhery2023palm} and LLaMA \cite{touvron2023llama, touvron2023llama2}. Following the success of pre-trained LLMs, researchers developed Multimodal LLMs (MLLMs) that can process cross-modality inputs. A specific, and particularly successful type is Vison-Language Foundation Models (VLMs), bridging visual information to text. VLMs are pre-trained on large-scale cross-modal datasets comprising images and languages \cite{clip,li2019visualbert,lu2019vilbert, wang2021simvlm,blip2,flamingo}. 

The integration of LLMs into the domain of autonomous driving \cite{jin2023adapt, wen2023dilu, wen2023road}, exemplified by methods such as GPT-Driver \cite{mao2023gpt} and DriveGPT4 \cite{xu2023drivegpt4}, demonstrate the potential for these models to leverage their pre-trained knowledge and reasoning abilities to interpret and predict vehicular dynamics and environmental interactions. 
VLMs have shown a unique strength in merging visual data with linguistic context, enabling a more holistic understanding of driving scenarios. 
Models like DriveMLM \cite{wang2023drivedreamer} utilize a multi-modal approach combining images, point clouds, and textual prompts to generate decision commands. 
This approach aids in the early identification of discrepancies or anomalies in predicted trajectories.

In this work, we focus on leveraging VLM to detect and explain hard cases that are difficult for trajectory prediction models, and distill high-level data from our pipeline to help effective model training.

\section{Methodology}
We evaluate our hypothesis in two stages, as illustrated in \cref{fig:arc}. In the first stage, we aim to verify the capability of VLM to detect hard cases in motion prediction.
We use results from existing prediction networks as a benchmark.
Unlike post-perception prediction networks that rely on detected bounding boxes, our approach focuses on prediction networks where inputs are derived directly from raw sensor data, such as images. 
As explained earlier, camera images provide richer visual information, aiding in the more effective detection of hard cases.

Given an input consisting of sequential image frames from multiple camera views, a motion prediction model generates predicted trajectories of different traffic agents. We then rank agents based on the Average Displacement Errors (ADE) between these predicted trajectories and the ground truth, from highest to lowest ADE.

The same sequential image frames are also fed into a VLM, alongside designed prompts where VLM is asked to (a) estimate a ranking of hard-to-predict agents, (b) assign an overall difficulty level to the scenario, and (c) provide explanations. 
In this paper, we utilize the online GPT-4v model \cite{openai2023gpt4} trained on extensive datasets.
As demonstrated in \cref{tab:user_example}, our prompts guide the VLM to process consecutive frames containing six-view camera images and a Bird-Eye-View (BEV) figure, with labeled agents including cars, trucks, pedestrians, cyclists, and motorcycles.
The VLM’s objectives include ranking these agents by prediction difficulty and scoring the scenario’s overall difficulty on a scale of 1 to 10. 
To enhance VLM's accuracy, two examples are provided as few-shot learning \cite{brown2020language}.
Finally, given the test images and descriptions, VLM is instructed to provide outputs in a certain format.
We assess the VLM's accuracy in mirroring the difficulty rankings of the existing prediction model using various ranking metrics, which we discuss later in \cref{sec:exp-metric}.

In the second stage, we explore the usability of the first stage using the data selection as a test example. 
Instead of training on the entire dataset of driving scenes, the VLM selects a subset of relatively difficult scenes based on their estimated difficulty levels. We evaluate the performance of networks trained on this smaller dataset against those trained on the full dataset.
The approach tests the potential of VLMs to improve training efficiency by creating a smaller, yet representative portion of the complete training dataset.

\section{Experiments}
\subsection{Datasets and Networks}
\label{sec:exp-dataset}
In our experiments, we evaluate our hypothesis together with GPT-4v using NuScenes dataset \cite{caesar2020nuscenes} for motion prediction from camera images. 
Given the past 3 frames of images (at $2~\si{Hz}$) from six different camera angles (front left, front, front right, back right, back, back left), the task is to predict the future 12 frames of detected agents. The detection module is an integrated part of the entire network.

We utilize two state-of-the-art methods including UniAD \cite{uniad} and ViP3D \cite{vip3d} to generate reference results for the agent ranking based on the high prediction error. UniAD, a planning-oriented method, incorporates modules for perception (detection, map segmentation, tracking),  prediction (for trajectory and occupancy), and planning. In our experiment, we focus on its perception and trajectory prediction modules. ViP3D is an end-to-end visual trajectory prediction method that includes detection, tracking, and prediction. For each sample, we first collect the predicted trajectories of agents successfully detected by both UniAD and ViP3D. The agent ranking order is then determined based on the average displacement error calculated against the ground truth trajectories.

Using the agent ranking from UniAD as ground truth, we compare it with several orders: a random order, the order from ViP3D, and GPT-4v. For the random ranking, we test on 10,000 trials to ensure statistical reliability and obtain a reference range for different metrics to our data.

\begin{table}
\def\arraystretch{1.3}
\caption{Ablation study on GPT-4v performance with different inputs and prompts. Larger values indicate a high correlation between GPT-4v ranking and UniAD ranking. \textit{Cam / Bev} means inputs of camera RGB images or bird-eye-view images with HD maps. \textit{Cam+Bev} indicates combining them as shown in \cref{tab:user_example}. \textit{Pos / Type} means if pixel positions of the labeled agent boxes / agent types are provided. IDs are used to distinguish different experiment settings.}
\begin{tabular}{c|c|cc|cccc} 
\toprule
ID & Input                                                                & Pos & Type & C-index        & NDCG           & \begin{tabular}[c]{@{}c@{}}Top 5\\Acc\end{tabular}  & \begin{tabular}[c]{@{}c@{}}Kendall's\\Tau\end{tabular}   \\ 
\midrule
1  & \multirow{2}{*}{Cam}                                                 &     & \checkmark  & 0.537          & 0.822          & 70.8\%          & 0.074           \\
2  &                                                                      & \checkmark & \checkmark  & 0.521          & 0.821          & 70.8\%          & 0.043           \\ 
\midrule
3  & Bev                                                                  & $-$   & \checkmark  & 0.522          & 0.830          & 68.1\%          & 0.044           \\ 
\midrule
4  & \multirow{4}{*}{\begin{tabular}[c]{@{}c@{}}Cam\\+\\Bev\end{tabular}} &     &      & 0.513          & 0.824          & 70.8\%          & 0.026           \\
5  &                                                                      & \checkmark &      & 0.507          & 0.821          & \textbf{72.2\%} & 0.015           \\
6  &                                                                      &     & \checkmark  & \textbf{0.570} & \textbf{0.837} & \textbf{72.2\%} & \textbf{0.140}  \\
7  &                                                                      & \checkmark & \checkmark  & 0.515          & 0.821          & 70.8\%          & 0.030           \\
\bottomrule
\end{tabular}
\label{tab:ablation}
\vspace{-1.0em}
\end{table}

\subsection{Evaluation Metrics} \label{sec:exp-metric}
To conduct an in-depth analysis of the estimated rankings, we employ four classic metrics for measuring the correlation of two rankings, briefly introduced below:
\subsubsection{Concordance Index (C-index)} C-Index \cite{harrell1996multivariable, steck2007ranking} is largely adopted in the medical domain, specifically in survival analysis \cite{DeepConvSurv}. It is defined as the ratio of the number of predicted concordant pairs, to the total number of pairs.
\subsubsection{Kendall's Tau} Similar to C-index, Kendall's Tau assesses ordinal association, calculated by subtracting the number of concordant pairs from the number of discordant pairs and then dividing by the total number of pairs.
\subsubsection{Top-K Accuracy} It calculates percentage of samples where the top $K$ predicted agents overlap completely with the top $K$ ground truths.
\subsubsection{Normalized Discounted Cumulative Gain (NDCG)} NDCG \cite{jarvelin2002cumulated} is common in Information Retrieval (IR) systems \cite{yilmazel2021intrinsic, qiao2019understanding}. Unlike the above three metrics, NDCG considers both ranking order and relevance scores, indicated by agents' minADE. If agents have similar minADEs, their ranking order affects NDCG less, making it more practical in our case.

\subsection{Quantitative Result of Agents' Rankings}
We visualize the results of hard-to-predict agents' ranking with different methods in \cref{fig:res_}.
It is observed that GPT-4V demonstrates relatively robust performance among all four metrics, achieving high scores. Based on the histogram of 10,000 random order distributions, it is clear that GPT-4V outperforms over 90\% of random cases.
Notably, the comparison between UniAD (ground truth) and ViP3D reveals an intriguing result: while there is no correlation (neither positive nor negative) in terms of C-index, Kendall's Tau, and Top-5 Accuracy, the consideration of the ADE correlation score as a ranking weight in NDCG places it at the highest score. 
For ViP3D and UniAD, although they are different models that might output the prediction with their own bias, their training on the same dataset may result in highly correlated outcomes.

\begin{figure}[t]
\centering
\includegraphics[width=\linewidth]{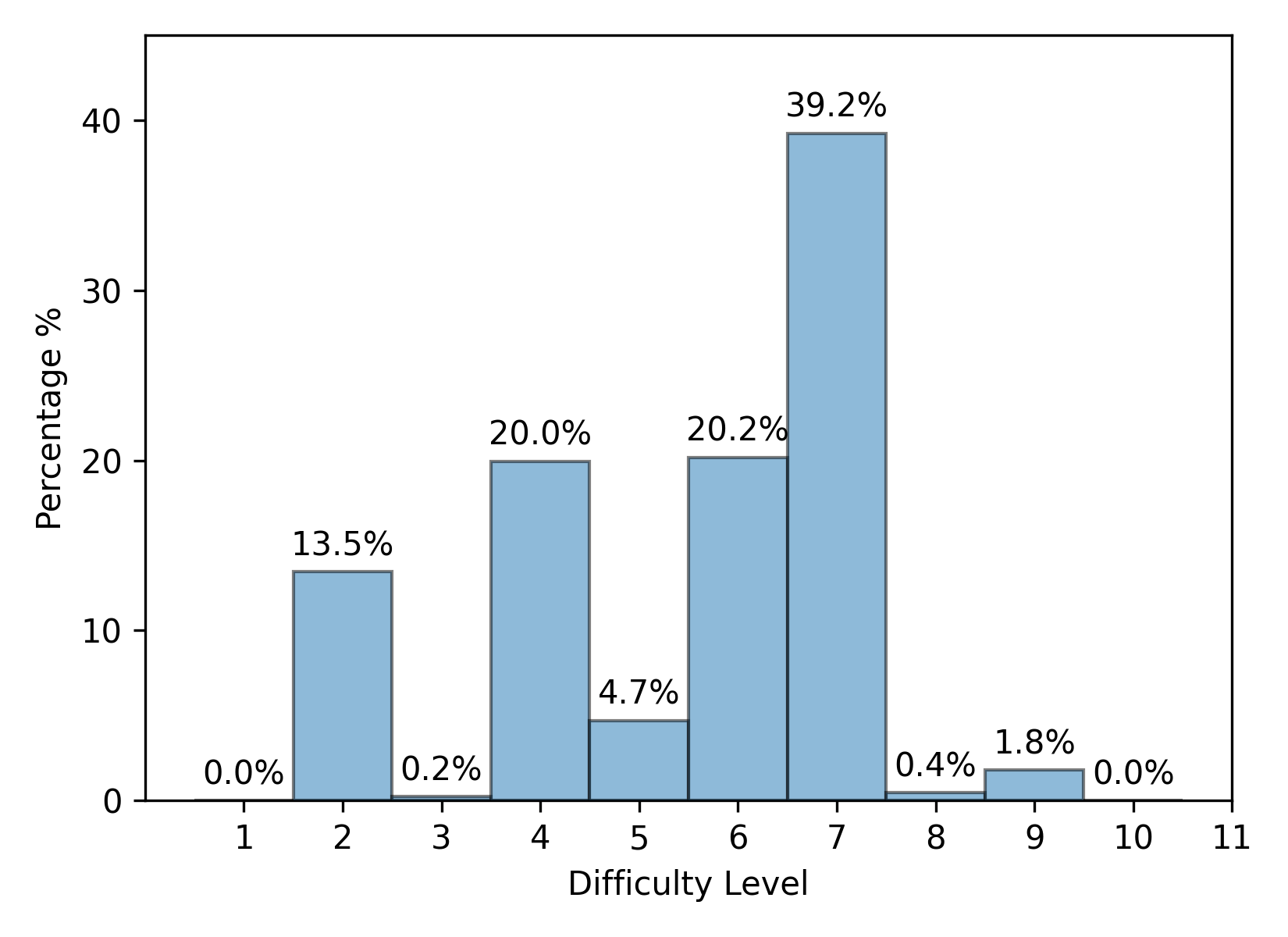}
\caption{Histogram of difficulty levels estimated by GPT-4v for 446 scenes.}
\label{fig:hist_diffuculty_level}
\vspace{-1.0em}
\end{figure}

\subsection{Ablation Study}
In our ablation study, we investigate the impact of varying user inputs and prompts on the outcomes generated by GPT-4v. The results in \cref{tab:ablation} highlight that experiment setting ID 6 stands out with the highest score.
This suggests that providing both camera and BEV images enhances the VLM's accuracy of the predicted rankings, aligning them more closely with the ground truth (UniAD). This is even more noticeable when we compare the experiment settings of ID 1, ID 3, and ID 6. Whereas comparing settings ID 4 and ID 6 shows that incorporating object type information into the inputs improves the GPT-4v's predictive accuracy. Interestingly, the comparison of ID 1 with ID 2, ID 4 with ID 5, and ID 6 with ID 7, indicates that adding numerical pixel positions (coordinates) of the labeled agents' boxes actually decreases the predictive accuracy.
This reduction in performance may suggest that the amount of few-shot learning provided is insufficient for GPT-4v to accurately interpret numerical pixel positions. Possibly, fine-tuning the model with a relative scale of data could enable GPT-4v to understand better and utilize these coordinates. 
Note that GPT-4v results shown in the \cref{fig:res_} are based on the settings used in ID 7.

\newcommand{\blue}[1]{$_{\color{TableBlue}\uparrow #1}$}
\begin{table}[t]
\caption{Comparison of minADE and minFDE (in meters) across various data selection settings trained with UniAD. We include results from models trained on the whole dataset, subsets selected randomly, and by GPT-4v. The \blue{blue} value represents the error percentage increase relative to the results from the whole dataset.}
\label{tab:ablation_study}
\def\arraystretch{1.3}
\label{tab:compare_sota}
    \centering
    \footnotesize
    \setlength{\tabcolsep}{3pt}
    \begin{adjustbox}{max width=\textwidth}
    \begin{tabular}{cc|cc|cc}
    \toprule
    \multicolumn{2}{c|}{Class}  & \multicolumn{2}{c|}{Vehicle} & \multicolumn{2}{c}{Pedestrain}  \\
    \midrule
    \multicolumn{2}{c|}{\# Samples {\scriptsize[Ratio\%]}}   & minADE & minFDE & minADE & minFDE \\ \midrule
    \parbox[t]{8mm}{\multirow{1}{*}{\rotatebox[origin=c]{0}{Whole}}}  & 28130 {\scriptsize[100]} & 0.71 & 1.02 & 0.82 & 1.11 \\ \midrule
    \parbox[t]{8mm}{\multirow{5}{*}{\rotatebox[origin=c]{0}{Random}}} & 2000 {\scriptsize[7.1]}  & 0.80\blue{13\%} & 1.22\blue{19\%} & 0.93\blue{13\%} & 1.31\blue{18\%} \\
    & 1000 {\scriptsize[3.6]} & 0.82\blue{16\%} & 1.27\blue{24\%} & 0.92\blue{12\%} & 1.29\blue{16\%}  \\
    & 500 {\scriptsize[1.8]}  & 0.93\blue{31\%} & 1.45\blue{42\%} & 0.99\blue{21\%} & 1.42\blue{28\%}  \\
    & 200 {\scriptsize[0.7]}  & 0.97\blue{36\%} & 1.55\blue{51\%} & 1.03\blue{25\%} & 1.45\blue{30\%}\\
    & 100 {\scriptsize[0.4]}  & 1.08\blue{52\%} & 1.73\blue{69\%} & 1.13\blue{38\%} & 1.63\blue{47\%}\\ \midrule
    \parbox[t]{8mm}{\multirow{2}{*}{\rotatebox[origin=c]{0}{GPT-4v}}}
    & 200 {\scriptsize[0.7]}  & 0.93\blue{31\%} & 1.48\blue{45\%} & 1.03\blue{25\%} & 1.43\blue{29\%} \\
    & 100 {\scriptsize[0.4]}  & 0.97\blue{37\%} & 1.56\blue{52\%} & 1.07\blue{31\%} & 1.56\blue{40\%}  \\
    \bottomrule
    \end{tabular}
    \end{adjustbox}
    \vspace{-1.0em}
\end{table}

\subsection{Data Selection}
Having confirmed GPT-4v's proficiency in predicting hard cases, we see many potential applications. As an example, we undertake a data selection task using the difficulty scores predicted by GPT-4v.

\begin{figure}[H]
  \begin{subfigure}{\linewidth}
    \centering
    \includegraphics[width=0.9\linewidth]{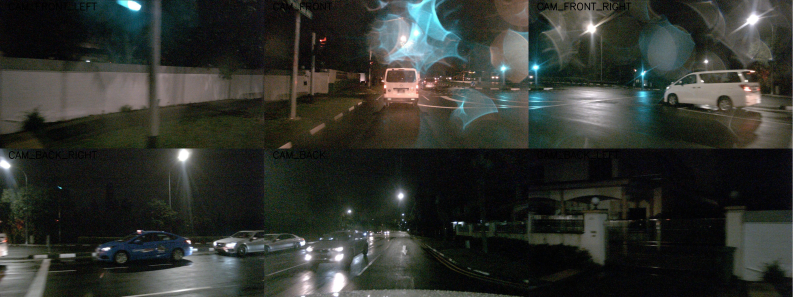}
    \caption{\fontsize{6.5pt}{6.5pt}\selectfont{GPT-4v: {\texttt{\color{RoyalBlue}Prediction difficulty is 9. \color{Black} 
The conditions are challenging due to nighttime driving and wet road surfaces, which can affect visibility and vehicle behavior. 
The reflections and glare from the lights can also make it difficult to accurately predict the movements of other road users. 
Additionally, the presence of multiple vehicles and a pedestrian crossing the street increases the complexity of the driving environment.
\vspace{1em}
}}}}

  \end{subfigure}
  \begin{subfigure}{\linewidth}
    \centering
    \includegraphics[width=0.9\linewidth]{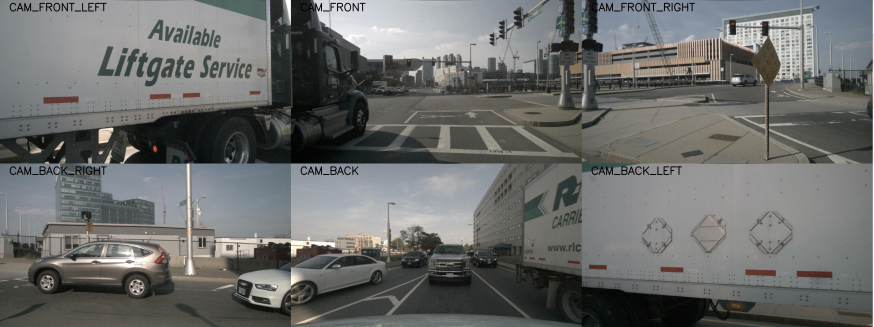}
    \caption{\fontsize{6.5pt}{6.5pt}\selectfont{GPT-4v:  {\texttt{\color{RoyalBlue}Prediction difficulty is 7. \color{Black}
    The vehicle is at an intersection with traffic lights and crosswalks, which adds complexity. 
There is a large truck on the left that may obstruct the view and movement, and there are multiple vehicles and pedestrians around. 
The presence of the truck and the busy intersection increase the difficulty of prediction due to potential blind spots and the need to anticipate the actions of several road users.
\vspace{1em}
}}}}
  \end{subfigure}
  \begin{subfigure}{\linewidth}
    \centering
    \includegraphics[width=0.9\linewidth]{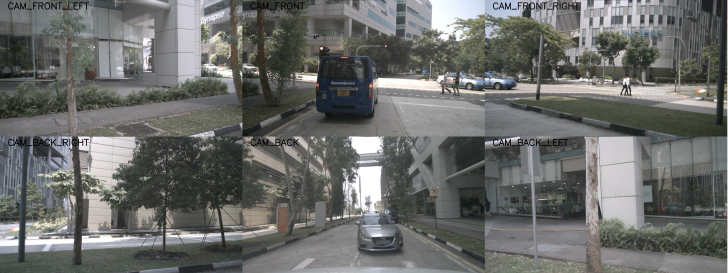}
    \caption{\fontsize{6.5pt}{6.5pt}\selectfont{GPT-4v:  {\texttt{\color{RoyalBlue}Prediction difficulty is 5. \color{Black} 
The vehicle is following a bus which might stop frequently to pick up and drop off passengers, adding some unpredictability. There are also pedestrians visible in the crosswalk, which requires careful monitoring. The intersection ahead adds complexity to the driving scenario, but overall traffic density is not high. The presence of the bus and pedestrians at the intersection moderately increases the difficulty of prediction.
\vspace{1em}
}}}}
  \end{subfigure}
  \begin{subfigure}{\linewidth}
    \centering
    \includegraphics[width=0.9\linewidth]{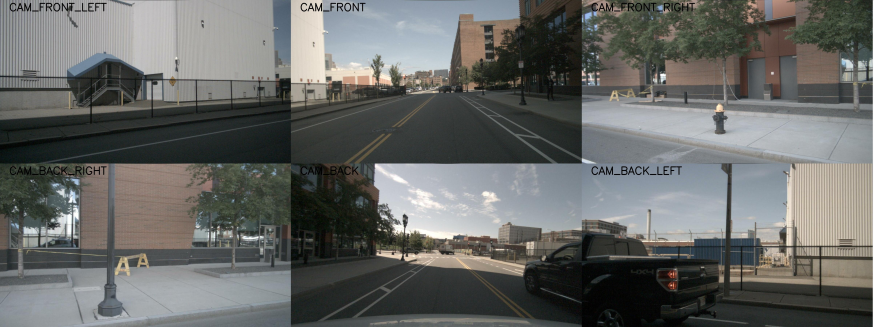}
    \caption{\fontsize{6.5pt}{6.5pt}\selectfont{GPT-4v:  {\texttt{\color{RoyalBlue}Prediction difficulty is 3. \color{Black}The traffic situation appears to be straightforward with light traffic and clear road markings. There are no immediate hazards or complex intersections that would increase the difficulty of prediction. The presence of a pedestrian on the sidewalk in the CAM FRONT RIGHT view does not significantly impact the prediction difficulty as they are not close to the roadway. The overall environment suggests a low level of complexity for driving prediction.}}}}
  \end{subfigure}

  \caption{GPT-4v scores the difficulty level for variable scenarios. Here we show four scenes scored as \{9, 7, 5, 3\} with explanations outputted by GPT-4v. A higher score means more difficulty.}
  \label{fig:vis_difficult_scenes}
\end{figure}

We begin by randomly selecting 451 training driving scenarios. GPT-4v then assesses each scenario, assigning a prediction difficulty score, ranging from 1 (easiest) to 10 (most difficult), with explanations. The assessment, guided by our prompts, is based on multiple factors, including road complexity, real-time traffic flow density, weather conditions, and any anomalous behavior or situation.
 The histogram in \cref{fig:hist_diffuculty_level} visualizes GPT-4v's score distribution. Note that 5 samples are missing in the histogram as there is a small chance that GPT-4v denies the output request claimed due to safety concerns. The figure indicates that over half of the scenarios are rated as not easy or challenging ($\geq$ 6), while very easy cases ($\leq$ 3) constitute less than 15\%. This score distribution also reflects that the NuScenes dataset already contains a high percentage of challenging scenarios. \cref{fig:vis_difficult_scenes} shows two visualization examples.
 
In our experiment, two training subsets, with 200 and 100 driving scenarios respectively, are selected from the high difficulty range (GPT-4v scores 7-9). As a baseline, we perform a random data selection, from the same 451 total driving scenes, to generate 200 and 100 driving scenarios. \cref{tab:ablation_study} shows the results. Note that we train on UniAD, more specifically, the second stage training for trajectory prediction. UniAD has two training stages: the first stage trains the perception part for initial BEV feature extraction weights, while the second stage involves training all modules. In our study, we utilize the weights from the first stage and continue the training with the selected data in the second stage, plugging out the original occupancy prediction and planning modules, as they are irrelevant to our objective.
To ensure fair comparisons, we align the training times across different setups to approximately 8 hours on 6$\times$A100.

The results in \cref{tab:ablation_study} show the performance improvement by data selection with GPT-4v when compared with randomly selected dataset. 
When the model is trained on 100 samples, which constitutes only the 0.4\% of the complete dataset, selected by GPT-4v, the minADE and minFDE increase by 37\% and 52\% for cars, and 31\% and 40\% for pedestrians, compared with the values trained on the whole dataset (28130 samples). These error values are comparable to those obtained with training on 200 randomly selected samples, double the training size, and much better than those obtained with training on 100 randomly selected samples, the same training size. Furthermore, the results obtained with training on 200 samples selected by GPT-4v are also better than those obtained with training on 200 randomly selected samples. Therefore, it can be observed that the training samples selected by GPT-4v, result in much better performance than the randomly selected samples.

\section{Conclusion}
In this paper, we propose a feasible pipeline that leverages VLM to detect hard cases in autonomous driving scenarios. Our work has verified the capability of GPT-4v in predicting hard cases, aligning with the existing prediction networks. We further explored the application of GPT-4v in the context of data selection, demonstrating that GPT-4v facilitates prediction networks to train efficiently via data selection.
Future work can be explored using offline VLM, where models and their pre-trained weights are open-sourced, allowing for closer integration with current autonomous driving systems. Furthermore, more potential work can be to resolve hard-to-predict cases incorporating VLM in the trajectory prediction pipelines. 

\section*{Acknowledgement}

This work\footnote{We have used ChatGPT for editing and polishing author-written text.} was funded by Vinnova, Sweden (research grant). The computations were enabled by the supercomputing resource Berzelius provided by National Supercomputer Centre at Linköping University and the Knut and Alice Wallenberg foundation, Sweden.

\bibliographystyle{IEEEtran}
\bibliography{IEEEabrv,ref}
\newpage
\subsection{Appendix}
\vspace{-1em}

\end{document}